\documentclass{article}
\usepackage{spconf,amsmath,graphicx}
\usepackage{multirow}
\usepackage{tipa}
\usepackage{url}

\title{Domain Adaptation For Formant Estimation Using Deep Learning}
\name
 {Yehoshua Dissen$^1$, Joseph Keshet$^1$\sthanks{This work was supported (in part) by grant from the MAGNET program of the Israeli Innovation Authority.}, Jacob Goldberger$^2$ and Cynthia Clopper$^3$}
\address{$^1$Department of Computer Science, Bar-Ilan University, Ramat-Gan, Israel\\
$^2$Engineering Faculty, Bar-Ilan University, Ramat-Gan, Israel\\
$^3$Department of Linguistics, Ohio State University, OH, USA}
\begin{document}
%
\maketitle
\begin{abstract}
In this paper we present a domain adaptation technique for formant estimation using a deep network. We first train a deep learning network on a small read speech dataset. We then freeze the parameters of the trained network and use several different datasets to train an adaptation layer that makes the obtained  network universal in the sense that it works well for a variety of speakers and speech domains with very different characteristics. We evaluated our adapted network on three datasets, each of which has different speaker characteristics and speech styles. The performance of our method compares favorably with alternative methods for formant estimation.
\end{abstract}
\section{Introduction}
\label{sec:intro}
Formant frequency estimation is among the most
fundamental problems in speech processing.  
In our previous work \cite{deepformants} we applied  deep learning methods 
 to the tasks of formant estimation and tracking. The network was trained and tested on the Vocal Tract Resonance  (VTR) database \cite{deng2006database} and  we achieved state-of-the art results. 
However, we experienced a dropoff in performance when the network was presented at test time with other recorded speech databases (e.g. \cite{clopper2014effects} \cite{hillenbrand1995acoustic}) due to the phenomenon of over-fitting to the speaker and speech domains represented in that database.
Our goal is to train a single network that performs well on a variety of different speaker and speech domains. A straightforward approach is to use at training phase not just the VTR dataset but also examples from all other available datasets.
However, when incorporating these databases into our training data, the results degraded for all domains of speech including the VTR dataset. These datasets were collected for different research purposes and there is a huge variability  among the datasets due to gender, age, dialects, recording conditions, etc.
Training a neural-network with a standard architecture based on fully connected layers that will yield good results on a diverse set of speaker domains is a very difficult task. Apart from formant estimation, the network neurons have to implicitly extract information from the input speech segment that describes the domain that contains this segment.  

In this paper we present a network architecture and a training procedure that  achieve accurate results regardless of the speech domain by introducing an additional domain adaptation layer that is trained separately from the original formant estimation network.  The training is performed  in two steps.
First, we train  a fully-connected network using the VTR dataset. Next, we freeze the network and, using all available  datasets, train an additional adaptation layer that transforms the network into  a formant estimation procedure that can be  automatically adapted to the domain of the input speech. We empirically demonstrate that both the proposed network architecture and the proposed training procedure are required to achieve state-of-the-art results on the three datasets used in the study.

The paper is organized as follows. The next section describes the speech features set.
Section \ref{sec:Datasets} describes the datasets used for training and evaluating the networks.
Section \ref{sec:Network} presents the domain adaptation deep network along with a training procedure. Section \ref{sec:Results} evaluates the proposed method by comparing it to state of the art formant estimation methods. We conclude the paper in Section \ref{sec:conclusion}.

\section{Acoustic Features}
\label{sec:format}

A key assumption is that in the task of formant estimation the whole segment is considered stationary, which mainly holds for monophthongs (pure vowels). The features are extracted from the whole segment.  

We use the same sets of features as in our previous work \cite{deepformants}. The first set is based on Linear Predictive Coding (LPC) analysis, while the second is based on the pitch-synchronous spectra. We briefly describe and motivate each set of features. 

\subsection{LPC-based features}

An LPC model determines the coefficients of a forward linear predictor by minimizing the prediction error in the least squares sense. The spectrum of the LPC model can be interpreted as the envelop of the speech spectrum. The model order $p$ determines how smooth the spectral envelop will be. Low values of $p$ represent the coarse properties of the spectrum, and as $p$ increases, more of the detailed properties are preserved. Beyond some value of $p$, the details of the spectrum do not reflect only the spectral resonances of the sound, but also the pitch and some noise. A disadvantage of this method is that if $p$ is not well chosen (i.e., to match the number of resonance present in the speech), then the resulted LPC spectrum is not as accurate as desired \cite{birch1988application}. 

Our first set of acoustic features are based on the LPC models with a range of model orders between 8 and 17. This way the classifier can combine or filter out information from different model resolutions. Practically, we use the LPC cepstral coefficients of order $30$  as  Atals method\cite{Atal74} was a better representation for deep learning classifiers than the LPC coefficients themselves \cite{deepformants}.

\subsection{Pitch-synchronous spectrum-based features}

The spectrum of a periodic speech signal is known to exhibit an impulse train structure located at multiples of the pitch period. A major concern when using the spectrum directly for locating the formants is that the resonance peaks might fall between two pitch lines, and then they are not ``visible''. The LPC model estimates the spectrum envelop to overcome this problem. Another method to estimate the spectrum while eliminating the pitch impulse train is using the \emph{pitch synchronous spectrum} \cite{MedanYair89}. According to this method the DFT is taken over frames the size of the instantaneous pitch. 

One of the main problem of this method is the need of a very accurate pitch estimator. In our previous work \cite{deepformants} we showed that using frames of the size of the median pitch along the segment produces a spectrum, where its picks are not contaminated by the pitch. 

\begin{figure}[t]
\includegraphics[width=0.4\textwidth]{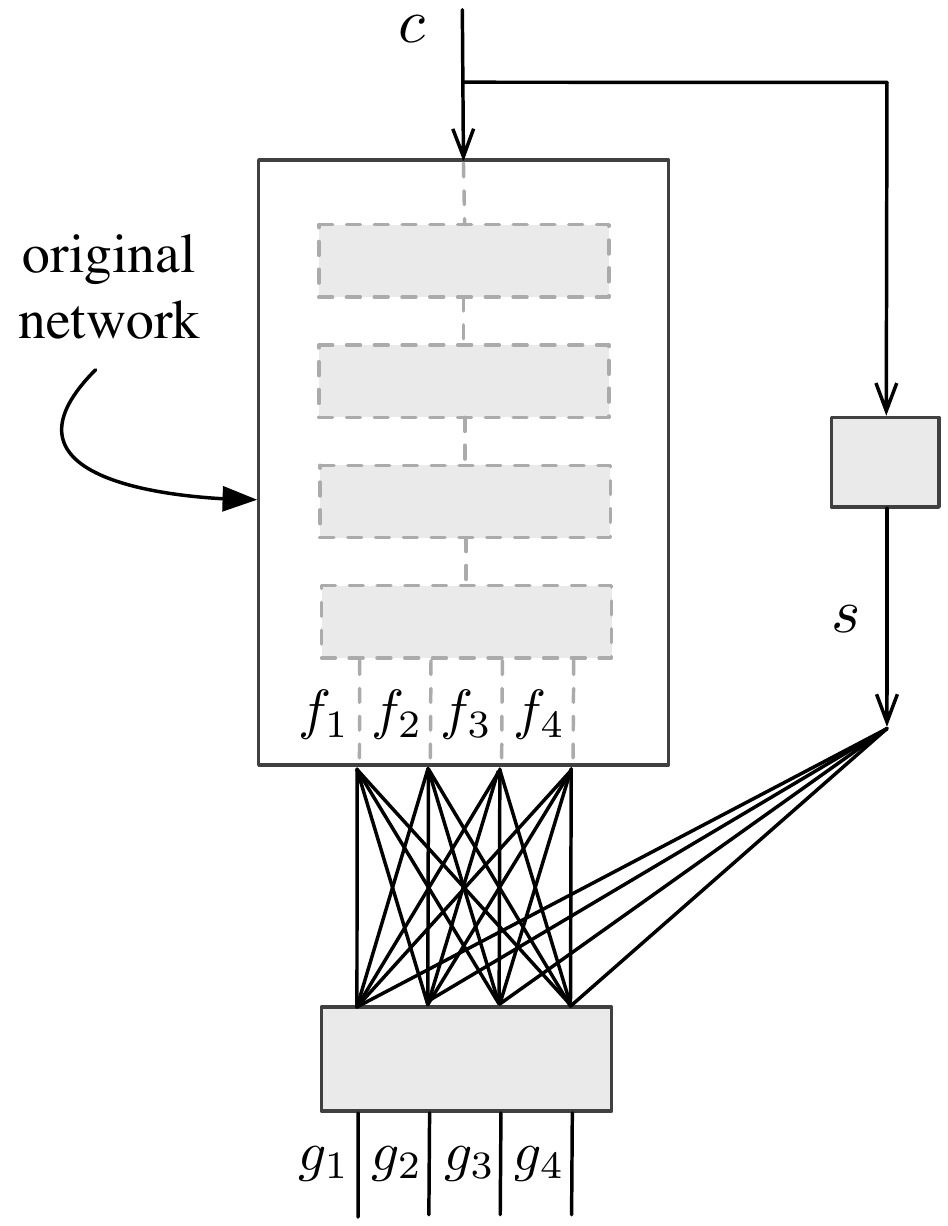}
\caption{A scheme of the domain adaptation formant estimation network. The network contains the VTR network component and the adaptation elements.}
\label{network_scheme}
\end{figure}
At the final stage, the resulting quasi pitch-synchronous spectrum is converted to cepstral coefficients by applying log compression and then Discrete Cosine transform (DCT). We use the first 50 DCT coefficients as our second set of features.

\section{Datasets}
\label{sec:Datasets}
 For the training and validating our model we used three different datasets.
 
The first being the Vocal Tract Resonance (VTR) corpus \cite{deng2006database}. This corpus is composed of 538 utterances selected as a representative subset of the well-known and widely-used TIMIT corpus. These were split into 346 utterances for the training set and 192 utterances for the test set. These utterances were manually annotated for the first 3 formants  and their bandwidths for every 10 msec frame. The fourth formant was annotated by the automatic tracking algorithm described in \cite{deng2004structured}, and it is not used here for evaluation. 

The second dataset \cite{clopper2014effects} contains segments of acoustic signal from 20 female native English speakers aged 18-22 with no history of speech or language deficits. The participants were evenly split between two American English dialects (Northern and Midland). Participants read aloud a list of 991 CVC words. This study focused on 39 target words (777 tokens) which did or did not have a lexical contrast between either /\textipa{E}/ vs. /\textipa{ae}/ (e.g., dead-dad vs. deaf-*daff) or /\textipa{A}/ vs. /\textipa{O}/ (e.g., cot-caught vs. dock-*dawk). We refer to this dataset as the \emph{Clopper} dataset. In this dataset the first and second formants were extracted using a standard tool (Praat) and then manually corrected.


The third dataset consists of data from a laboratory study conducted by Hillenbrand \cite{hillenbrand1995acoustic}. It contains segments of acoustic signal from 45 men, 48 women, and 46 ten-to 12-year-old children (27 boys and 19 girls). 87\% of the participants were raised in Michigan, primarily in the southeastern and southwestern parts of the state.  The audio recordings contain 12 different vowels (/\textipa{i, I, E, ae, A, O, U, u, 2, 3\textrhoticity, e, o}/) from the words: heed, hid, head, had, hod, hawed, hood, who'd, hud, heard, hayed, hoed. We refer to this dataset as the \emph{Hillenbrand} dataset.
As in the Clopper dataset,  formants were extracted using an LPC-based automatic tool and then manually corrected.

The Clopper and Hillenbrand datasets are very different from the VTR dataset: they include young females and children who have a much higher range of formants than average men and women in the VTR dataset.

\begin{figure*}[thp]
\centering
\includegraphics[width=.33\textwidth]{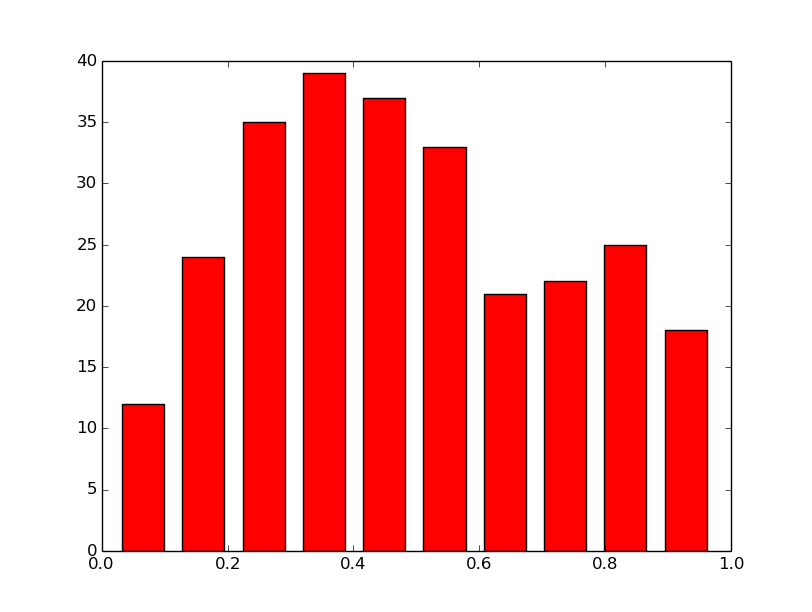}
\includegraphics[width=.33\textwidth]{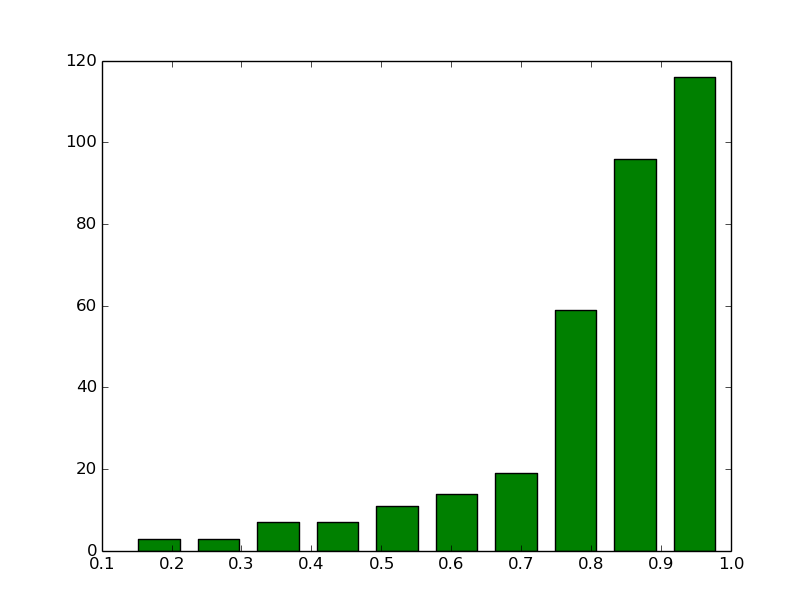}
\includegraphics[width=.33\textwidth]{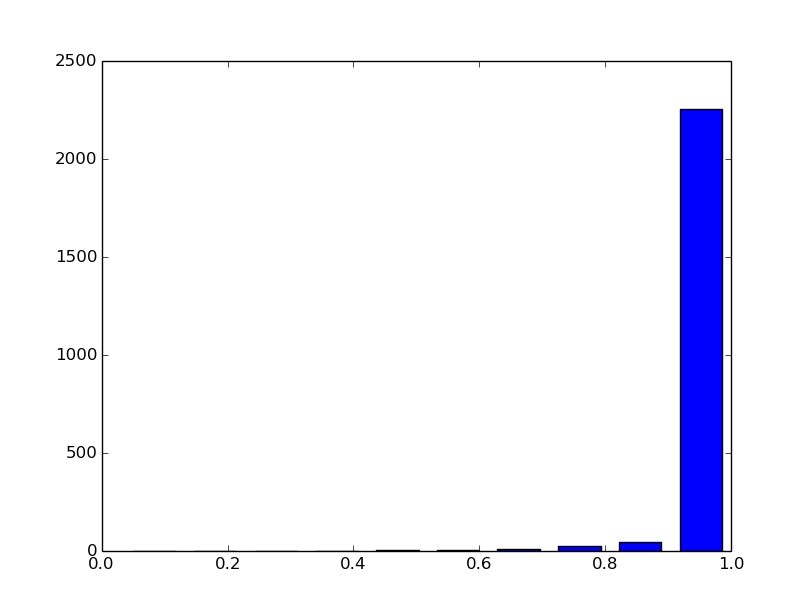}
\caption{Histograms of domain bias activations for the datasets Clopper (left), Hillenbrand (middle) and VTR (right).}
\label{fig:histograms}
\end{figure*}

\section{A Domain Adaptation Network}
\label{sec:Network}

In this section we describe a network architecture and  training procedure that jointly form a domain adaptation approach for formant estimation. 
The network used for formant estimation is a feed-forward network trained in two phases. First we train a network using only the VTR dataset which is small, with a limited set of speakers and the formant annotation is relatively reliable. The network consists of an input layer with 350 features, 3 fully connected hidden layers of 1024, 512 and 256 neurons respectively, and an output linear layer that provides an estimation of the 4 formants. 
Since the network is trained solely on the VTR dataset we denote it hereafter as the VTR network. 
The trained VTR network yields good results on the test subset of the VTR dataset. However when it is applied to other datasets there is a significant performance degradation.
The goal of this study is to train a single network that can be successfully used on different datasets. Naturally we do not assume that at train or test time the domain identity
of a given input is revealed to the network rather we want the network to learn to adapt its predictions based on the original features.
The approach we take in this study is to transfer the VTR network originally trained on the VTR dataset into a Domain Adaptation  (DA) network. This network adaptation is done by adding two new elements to the VTR network. First we use the VTR formant estimation output layer as input to another linear layer that produces a new formant estimation. We also add a  selection neuron $s$ whose input is the same input of the original network. This selection neuron controls the amount of network adaptation that should be applied to each  given input. 
The control element $s$ is a non-linear function of the input feature vector $c$ and is implemented by a linear operation followed by a sigmoid activation function, i.e.:
\begin{equation}
s(c) = \sigma(w_s \cdot c + b_s) 
\label{adapt2_eq}
\end{equation}
such that $c$ is the input vector and  $w_s$ and $b_s$ are network parameters that are learned at training phase.
Denote the formant output of the original VTR network by $f_1,...,f_4$.
The new formant estimation $g_1,...,g_4$ is computed as follows:
\begin{equation}
g_i = \sum_{j=1}^4 w_{ij} f_j + b_i + v_i \cdot  s (c)  
\label{adapt_eq}
\end{equation}
where $w_{ij}$ and $b_i$ are the parameters of the additional linear layer and $v_i$ is a multiplicative term that defines the contribution of the dataset control element $s(c)$ to the estimation of the $i$-th formant.  
A scheme of the domain adaptation (DA) network is shown in Figure \ref{network_scheme}.

We can use all the available datasets to train the DA network from scratch starting from a random initialization of the network parameters. However, since the network topology is complicated, this yields inferior results  (as we show in the next section) . Instead, we train the network in two steps. First we train the VTR sub-network only using only  the VTR dataset. Next we freeze the parameters of the VTR network and only train the adaptation parameters (\ref{adapt_eq}) (\ref{adapt2_eq}) using  the VTR, Clopper, and Hillenbrand datasets. This two-step training procedure ensures that the VTR sub-network is responsible for the core formant estimation, and the adaptation part of the network is responsible solely for adapting the formant estimation to the conditions of the specific input vector.

One of the advantages of training the network in two steps, is that  there is only a small number of adaptation parameters so large amounts of data from each dataset are not necessary. Hence, domains with limited labeled data as in our case can still be used to learn a good formant estimator.

To better visualize what the network has learned we show the histograms of the domain parameters activation values $s$ for each dataset. The histogram axis are the number of examples in each of the 10 buckets of activations between 0 and 1 (which is the output range of a sigmoid function).
As seen in the right-most histograms in Figure \ref{fig:histograms}, the $s$ values for the VTR database are almost exclusively concentrated in the same area showing that the network automatically found that no adaptation is needed for data from the VTR dataset. This coincides with the fact that the original network was trained on VTR dataset. In contrast, predictions from the Hillenbrand dataset needed to be corrected occasionally and predictions from the Clopper dataset consistently needed to be corrected, as seen by the variance in $s$ values.

\begin{table}[ht!]
\caption{\label{tab:results} {\it Estimation of formant frequencies using deep learning with and without domain adaptation  and compared to WaveSurfer. Boldface indicates the best result in that category.}}
\vspace{2mm}
\centering
\begin{tabular}{llccc}
\hline
\hline
 Dataset  & Method & $F_1$ & $F_2$ & $F_3$ \\ 
 \hline
 \multirow{3}{*}{VTR}
  & WaveSurfer \cite{sjolander2000wavesurfer}  & 70 & 96 &  154 \\
 & DeepFormants \cite{deepformants} & \bf{48} & \bf{83} &109  \\
 & Domain Adaptation   & 50 & 86 &  \bf{104} \\

 \hline
\multirow{3}{*}{Hillenbrand}
 & WaveSurfer & 68 & 190 &  182\\
& DeepFormants & 71 & 160 & 131  \\
 &Domain Adaptation  & \bf{36} & \bf{100} & 116\\

 \hline
\multirow{3}{*}{Clopper}
 & Wavesurfer & 128 & 181 &  -- \\
& DeepFormants & 228 & 168 & -- \\
 & Domain Adaptation  &  \bf{103} & \bf{157} & --\\
\hline
\hline
\end{tabular}
\end{table}

\section{Results}
\label{sec:Results}
Here we present the results of the domain adaptation network compared to our previous work \cite{deepformants} that is based on training only on the VTR data and WaveSurfer \cite{sjolander2000wavesurfer} a popular tool in phonetic research.
The results of our system, DeepFormant and of WaveSurfer on the three datasets we used are shown in Table \ref{tab:results}, where the loss is the mean absolute difference in Hz. Note that the Clopper dataset was annotated for the first and the second formants, hence the third formant was not evaluated.

As seen in the table, we achieved better results across the board over WaveSurfer when comparing our respective estimations to the manually annotated reference. The domain adaptation network shows improvement over DeepFormants in both the Clopper and Hillenbrand datasets with no significant drop off in accuracy on the VTR dataset. These results show the advantage of the proposed network architecture over standard networks based on fully connected layers.

When comparing these results to separate networks trained and tested on each of the databases, i.e. training a model with data from the Clopper dataset and then testing on the Clopper test set and another model trained and tested on the Hillenbrand dataset in the same manner and so on for the VTR dataset, we obtained comparable results. Hence,  there is no need for multiple models for each domain, this single network can separate between the speaker and speech domains and adjust its estimations accordingly.

In the next experiment we demonstrate the need for the two step training procedure proposed in this study.
Table \ref{tab:together} shows the formant estimation results of two training strategies of the domain adaptation network. In the table we compare the results of the two step training to the results of a network with the same topology but trained jointly on all three datasets. As can be seen from Table \ref{tab:together},  other than for the first formant in the Clopper dataset the accuracy is greatly diminished across all other data. Moreover adding the selection layer doesn't improve results over training a network identical to the VTR network but using all three datasets during training.
\begin{table}[ht!]
\caption{\label{tab:together} {\it Results of formant estimation for two possible training procedures of the proposed domain adaptation network. Boldface indicates the best result in that category.}}
\vspace{2mm}
\centering
\begin{tabular}{llccc}
\hline
\hline
Dataset & training method & $F_1$ & $F_2$ & $F_3$ \\ 
 \hline
\multirow{2}{*}{VTR} & joint training & 96 & 279 &  283 \\ 
&  two step training & \bf{50} & \bf{86} &  \bf{104}\\ 
\hline
\multirow{2}{*}{Hillenbrand} & joint training & 71 & 119 & 129\\ 
& two step training  & \bf{36} & \bf{100} & \bf{116} \\ 
\hline
\multirow{2}{*}{Clopper} & joint training  & \bf{70} & 166 &  -- \\
& two step training & 103 & \bf{157} & --\\ 
\hline
\hline
\end{tabular}
\end{table}

\section{Conclusions}\label{sec:conclusion} 

In this paper we proposed a formant estimation deep-learning architecture  that achieves state-of-the-art results across several speech and speaker domains that are very different in nature. 
We also proposed a training scheme that validates the claim that each component of the network is indeed responsible for the task it was designed to do, either formant estimation or domain adaptation.
We have demonstrated here automated formant
estimation tools that are ready to be added to the
methods that sociolinguists use to analyze acoustic data. The
tools will be publicly available at \url{https://github.com/
MLSpeech/domainadaptation}.
In future work we  plan to evaluate the robustness of our method not only to different datasets but also to noisy environments.

\bibliographystyle{IEEEbib}
\bibliography{strings}

\begin{thebibliography}{1}

\bibitem{deepformants}
Yehoshua Dissen and Joseph Keshet,
\newblock ``Formant estimation and tracking using deep learning,''
\newblock {\em The 17th Annual Conference of the International Speech
  Communication Association}, 2016.

\bibitem{deng2006database}
Li~Deng, Xiaodong Cui, Robert Pruvenok, Yanyi Chen, Safiyy Momen, and Abeer
  Alwan,
\newblock ``A database of vocal tract resonance trajectories for research in
  speech processing,''
\newblock in {\em Acoustics, Speech and Signal Processing, 2006. ICASSP 2006
  Proceedings. 2006 IEEE International Conference on}. IEEE, 2006, vol.~1, pp.
  I--I.

\bibitem{clopper2014effects}
Cynthia~G Clopper and Terrin~N Tamati,
\newblock ``Effects of local lexical competition and regional dialect on vowel
  production,''
\newblock {\em The Journal of the Acoustical Society of America}, vol. 136, no.
  1, pp. 1--4, 2014.

\bibitem{hillenbrand1995acoustic}
James Hillenbrand, Laura~A Getty, Michael~J Clark, and Kimberlee Wheeler,
\newblock ``Acoustic characteristics of american english vowels,''
\newblock {\em The Journal of the Acoustical society of America}, vol. 97, no.
  5, pp. 3099--3111, 1995.

\bibitem{birch1988application}
Gary~E Birch, Peter Lawrence, John~C Lind, and Robert~D Hare,
\newblock ``Application of prewhitening to ar spectral estimation of {EEG},''
\newblock {\em Biomedical Engineering, IEEE Transactions on}, vol. 35, no. 8,
  pp. 640--645, 1988.

\bibitem{Atal74}
Bishnu~S Atal,
\newblock ``Effectiveness of linear prediction characteristics of the speech
  wave for automatic speaker identification and verification,''
\newblock {\em the Journal of the Acoustical Society of America}, vol. 55, no.
  6, pp. 1304--1312, 1974.

\bibitem{MedanYair89}
Yoav Medan and Eyal Yair,
\newblock ``Pitch synchronous spectral analysis scheme for voiced speech,''
\newblock {\em IEEE Trans. on Acoustics, Speech and Signal Processing}, vol.
  37, no. 9, pp. 1321--1328, 1989.

\bibitem{deng2004structured}
Li~Deng, Leo~J Lee, Hagai Attias, and Alex Acero,
\newblock ``A structured speech model with continuous hidden dynamics and
  prediction-residual training for tracking vocal tract resonances,''
\newblock in {\em Acoustics, Speech, and Signal Processing, 2004.
  Proceedings.(ICASSP'04). IEEE International Conference on}. IEEE, 2004,
  vol.~1, pp. I--557.

\bibitem{sjolander2000wavesurfer}
K{\aa}re Sj{\"o}lander and Jonas Beskow,
\newblock ``Wavesurfer-an open source speech tool.,''
\newblock in {\em Interspeech}, 2000, pp. 464--467.

\end{thebibliography}

\end{document}